\documentclass[11pt]{article}

\usepackage[preprint]{acl}

\usepackage{times}
\usepackage{latexsym}
\usepackage{makecell}
\usepackage{tabularx}
\usepackage[T1]{fontenc}

\usepackage[utf8]{inputenc}

\usepackage{microtype}

\usepackage{inconsolata}

\usepackage{amsmath,graphicx,hyperref}
\usepackage{booktabs,multirow}
\usepackage{color}
\usepackage[table]{xcolor}
\usepackage{amsfonts}
\usepackage{amssymb}
\usepackage{xspace}
\usepackage{svg}
\usepackage{listings}
\usepackage{tcolorbox}
\usepackage{caption}
\usepackage{tcolorbox}
\usepackage{multicol}
\usepackage{cuted} 
\newcommand{\sys}{$\mathsf{SciAux}$\xspace}

\title{Thinking in a Crowd: How Auxiliary Information Shapes LLM Reasoning}

\author{Haodong Zhao\thanks{Equal contribution.}, Chenyan Zhao$^{*}$, Yansi Li, Zhuosheng Zhang$^{\dagger}$, Gongshen Liu\thanks{Corresponding authors.} \\
        School of Computer Science, Shanghai Jiao Tong University\\
 \small{
   \texttt{\{zhaohaodong, zhao.ccc, yansi\_li, zhangzs, lgshen\}@sjtu.edu.cn}} \\
}

\begin{document}
\maketitle
\begin{abstract}
The capacity of Large Language Models (LLMs) to reason is fundamental to their application in complex, knowledge-intensive domains. In real-world scenarios, LLMs are often augmented with external information that can be helpful, irrelevant, or even misleading. This paper investigates the causal impact of such auxiliary information on the reasoning process of LLMs with explicit step-by-step thinking capabilities. We introduce \sys, a new dataset derived from ScienceQA, to systematically test the robustness of the model against these types of information. Our findings reveal a critical vulnerability: the model's deliberative ``thinking mode'' is a double-edged sword. While helpful context improves accuracy, misleading information causes a catastrophic drop in performance, which is amplified by the thinking process. Instead of conferring robustness, thinking reinforces the degree of error when provided with misinformation. This highlights that the challenge is not merely to make models ``think'', but to endow them with the critical faculty to evaluate the information upon which their reasoning is based.
The \sys dataset is available at \href{https://huggingface.co/datasets/billhdzhao/SciAux}{https://huggingface.co/datasets/billhdzhao/SciAux}.
\end{abstract}

\section{Introduction}
\label{sec:intro}

Recent advancements in Large Language Models (LLMs) have pushed their capabilities beyond simple pattern matching and information retrieval into the realm of complex reasoning. This evolution has enabled their deployment in high-stakes, knowledge-intensive fields where the quality of reasoning is paramount. A dominant paradigm for improving LLM factuality and providing up-to-date knowledge is Retrieval-Augmented Generation (RAG) and tool usage, which couples an LLM with external knowledge~\cite{lewis2020retrieval, gao2023retrieval}. However, this paradigm introduces a critical dependency: Performance is dependent on the quality of the information retrieved. This presents an unforeseen consequence, as the very feature designed to ground models in reality also introduces a new vector for failure.

The Achilles heel of this approach is the ``garbage in, garbage out'' problem. Real-world retrieval systems are imperfect, often returning a noisy mixture of highly relevant, tangentially related, and sometimes factually incorrect or misleading documents~\cite{wu2024easily}. This informational landscape presents a fundamental challenge to the robustness of LLM reasoning. An LLM must not only comprehend the retrieved context, but also critically evaluate its relevance and veracity to answer a given query correctly. This requires a deeper understanding of how LLMs process, reason, and are influenced by the auxiliary information they receive~\cite{nye2021show}.
\begin{figure}[t]
	\centering
	\includegraphics[width=1\linewidth]{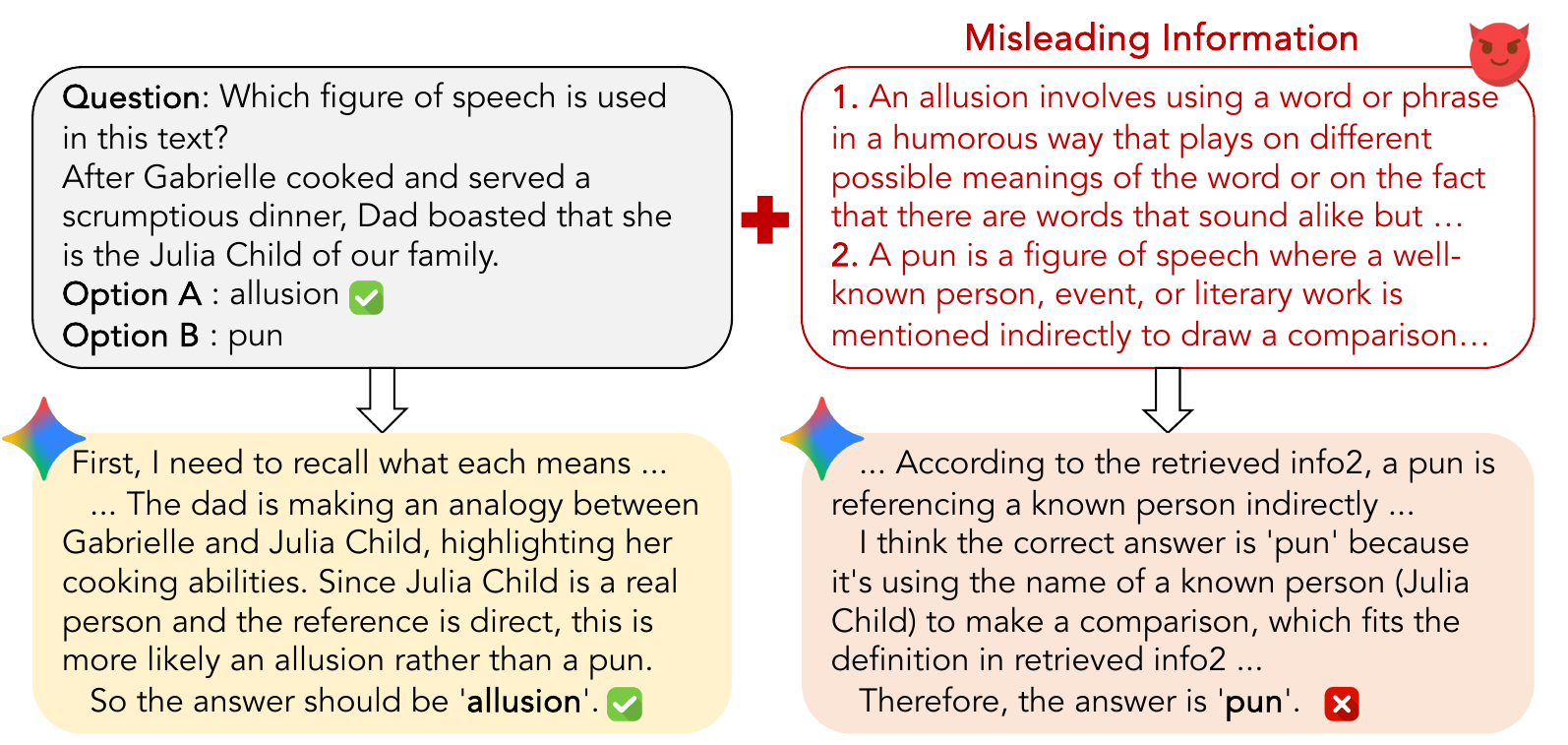}
	\caption{Examples where reasoning with the auxiliary information resulted in failure.}
	\label{fig:demo}
\end{figure}

Much of the previous research on context distraction~\cite{shi2023large} and misinformation~\cite{zou2023representation} has treated the LLM as a monolithic black-box operation, focusing on input-output behaviors without interrogating the reasoning process itself. This work moves beyond that limitation by leveraging Large Reasoning Models (LRMs) that feature an explicit, switchable ``thinking mode'' designed for deliberate step-by-step reasoning on complex tasks~\cite{yang2025qwen3,li2025dancing}. This architectural feature provides a unique opportunity for a controlled, causal investigation into the reasoning process, allowing us to ask a more refined set of questions about the fundamental trade-offs and failure modes of deliberative AI cognition.

This leads to the central question this paper seeks to answer: Does an reasoning process enhance an LLM's ability to discern and reject bad information? Or, paradoxically, does it deepen the model's vulnerability by forcing it to engage with and rationalize flawed premises? The development of ``thinking'' models is intended to overcome the limitations of intuitive systems, yet this very advancement may introduce a more subtle and potentially more dangerous failure mode. To investigate this question, we first construct a dataset \sys where each question is provided with different types of auxiliary information. We study the performance of latest LRMs with switchable thinking modes when provided with different information and in different modes, demonstrating the double-edged sword effect of the reasoning model's thinking mode.
This paper makes the following contributions to the understanding of LLM reasoning:

$\bullet$ We introduce the \sys dataset, which is designed to evaluate LLM reasoning under controlled informational stress by providing a set of helpful, irrelevant and misleading context snippets.\looseness=-1

$\bullet$ Our empirical analysis demonstrating that while reasoning is beneficial in clean or helpful contexts, it becomes a double-edged sword in the presence of misleading information.

$\bullet$ We propose an explainable classification of failures caused by the reasoning process and reveal the main failure causes of different models. We hope that this work will inspire the design of subsequent reasoning models.
\section{Related Work}
\label{sec:related}

\subsection{CoT and Reasoning in LLMs}
Chain-of-Thought (CoT) prompting, which directs models to produce step-by-step reasoning, significantly improves performance on complex reasoning tasks~\cite{wei2022chain}. This methodology has led to various extensions and its integration into LRMs~\cite{shojaee2025illusion}, such as the Qwen3 series that supports dynamic switching between intuitive and deliberative reasoning modes~\cite{yang2025qwen3}. However, assessing the fidelity and logical soundness of explicit reasoning traces remains challenging, as correct final answers do not necessarily indicate robust intermediate reasoning~\cite{lee2025evaluating}.

\subsection{Robustness and Knowledge Grounding in LLMs}
LLMs exhibit significant vulnerability to irrelevant and misleading contextual information, with even subtle distractions leading to notable declines in performance~\cite{shi2023large, wu2024easily, cavusoglu2024disgem, zou2023representation, wallace2019universal}. This issue is particularly pronounced in Retrieval-Augmented Generation (RAG) systems, which ground outputs in external knowledge to mitigate hallucinations~\cite{lewis2020retrieval, gao2023retrieval}. However, the overall effectiveness of RAG is fundamentally constrained by the quality of retrieved content, making robustness to informational noise a critical requirement~\cite{khattab2020colbert,asai2024self}. Motivated by these challenges, our \sys benchmark systematically evaluates LLM reasoning under both irrelevant and misleading auxiliary information.

\section{The \sys Dataset}
\label{sec:dataset}

To systematically investigate the impact of auxiliary information on LLM reasoning, we introduce the \sys dataset. The dataset is constructed by augmenting questions from the well-established ScienceQA dataset~\cite{lu2022learn}.

\subsection{Foundation: The ScienceQA Dataset}
When trying to find a dataset for simple validation of model inference, most question-answering (QA) datasets~\cite{kwiatkowski-etal-2019-natural,mallen2023trustlanguagemodelsinvestigating, krithara2023bioasq} generally suffer from the issue that most questions are knowledge-based, relying on factual recall rather than reasoning.
ScienceQA~\cite{lu2022learn} has broad coverage of scientific domains, and emphasis on multi-hop reasoning, making it ideal for assessing complex cognitive tasks. Importantly, it provides human-annotated \texttt{lecture} and \texttt{solution} fields: the \texttt{lecture} offers relevant background knowledge, while the \texttt{solution} presents a step-by-step explanation. These annotations not only establish a gold standard for helpful context but also facilitate the generation of misleading context variants.

When constructing \sys, we exclude the questions in ScienceQA that require images to answer, those with incomplete statements in the \texttt{question} field, and those that do not include the \texttt{lecture} field. In total, we select 6,828 entries to build the \sys dataset.

\subsection{Generate Auxiliary Information}
For each selected question, we generate a corresponding set of six auxiliary information snippets, two each of which are helpful, irrelevant, or misleading.
Specifically, we use \texttt{Gemini 2.5 Pro} to generate the snippets. In constructing the prompt, we provide Gemini with the \texttt{question}, the \texttt{choices}, and the \texttt{lecture} field. We explicitly instruct Gemini to generate six types of auxiliary information snippets based on these inputs:

$\bullet$ \textit{Helpful information (H1 and H2):} two correct pieces of information that help answer the question.\looseness=-1

$\bullet$ \textit{Misleading information (M1 and M2):} two pieces of information related to the topic of the question but contradict the provided lecture.\looseness=-1

$\bullet$ \textit{Irrelevant information (I1 and I2):} two pieces of information that are related to the topic of the question but not relevant to answering the question.

In addition, we emphasize that each piece of information should be provided as an independent background knowledge item, and the six auxiliary information to be as similar as possible in length in the prompt. We give the prompt template in Section~\ref{sec:apd-prompt}, and Table~\ref{tab:sciaux_example} gives a sample from \sys.

\subsection{Dataset Analysis and Statistics}
\begin{figure}[h]
	\centering
	\includegraphics[width=0.95\linewidth]{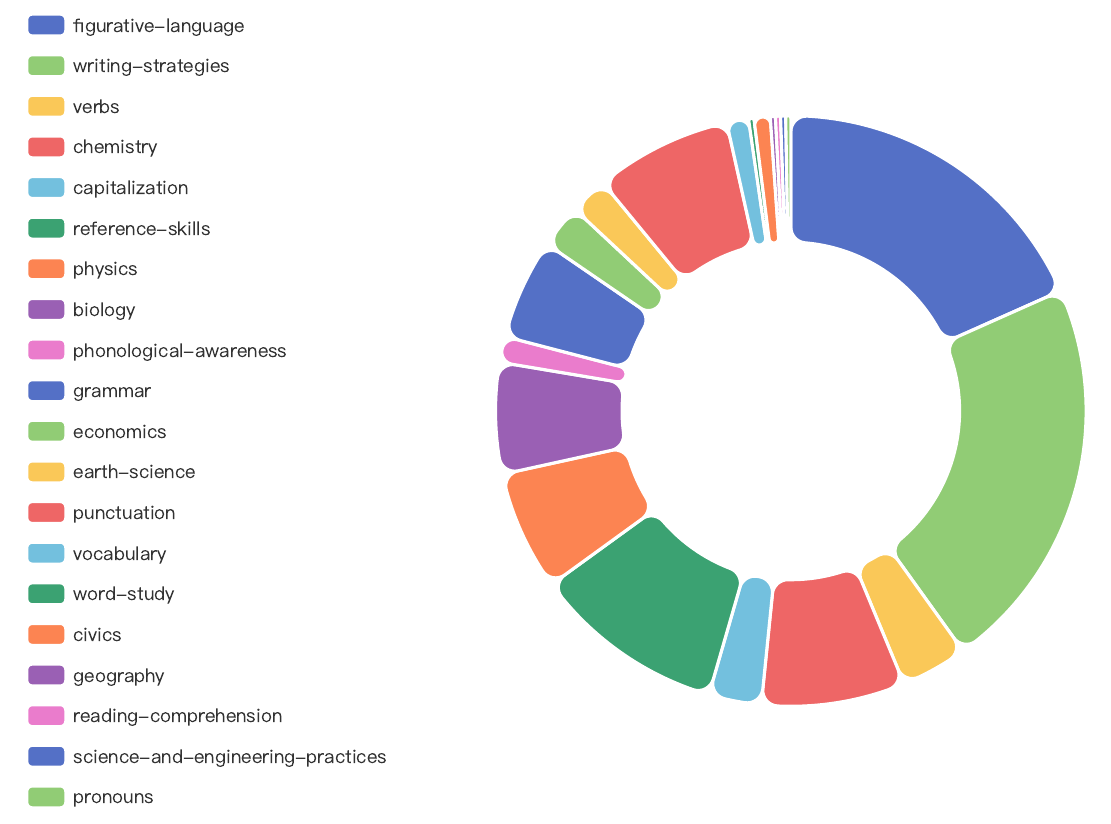}
	\caption{Topic distribution of the \sys dataset.}
	\label{fig:datasetTopic}
\end{figure}

\begin{table*}[h]
\centering
\resizebox{\textwidth}{!}{
\begin{tabular}{@{}lp{1.88\columnwidth}@{}}
\toprule
\textbf{Component} & \textbf{Content} \\ \midrule
\textbf{ID} & 62 \\
\textbf{Question} & Which figure of speech is used in this text?

Rebecca's Bistro used to be a great place to go for a delicious and carefully prepared dinner, but nobody goes there anymore: it's too crowded. \\
\textbf{Choices} & (0) apostrophe, (1) paradox\\
\textbf{Answer} & (1) \\ \midrule
\textbf{H1} & A paradox is a figure of speech presenting a statement or situation that appears self-contradictory or logically impossible on the surface. However, upon reflection, it often reveals an unexpected or deeper truth, challenging conventional assumptions. \\
\textbf{H2} & Apostrophe, as a figure of speech, occurs when a speaker or writer directly addresses someone who is absent, deceased, or imaginary. It can also involve addressing an inanimate object or an abstract concept as if it were a person capable of responding. \\
\textbf{M1} & A paradox is simply a statement that uses exaggeration for emphasis or effect, making a claim that is not meant to be taken literally. For example, saying 'I'm so hungry I could eat a horse' is a classic use of paradox in everyday language. \\
\textbf{M2} & Apostrophe refers to the grammatical mark (') used primarily to indicate possession (e.g., 'Rebecca's Bistro') or the omission of letters in contractions (e.g., 'it's'). Its function is purely grammatical, not related to addressing entities.\\
\textbf{I1} & "Hyperbole is a figure of speech characterized by deliberate and obvious exaggeration for emphasis or humorous effect, not intended to be taken literally. An example is 'This bag weighs a ton,' describing something very heavy. \\
\textbf{I2} & Oxymoron is a figure of speech that combines two seemingly contradictory or opposite terms to create a striking effect. Examples include 'jumbo shrimp,' 'deafening silence,' or 'bittersweet,' highlighting complexity or irony. \\ \bottomrule
\end{tabular}
}
\caption{An Example from the \sys Dataset.}
\label{tab:sciaux_example}
\end{table*}
In total, the \sys dataset contains 6,828 items. Figure~\ref{fig:datasetTopic} presents the distribution of the topics covered in the dataset. In general, the dataset covers 20 different topics, including writing strategies, figurative language, biology, physics, and other topics. This diversity ensures that \sys is not biased toward a narrow subject area.

\begin{figure}[h]
	\centering
	\includegraphics[width=0.8\linewidth]{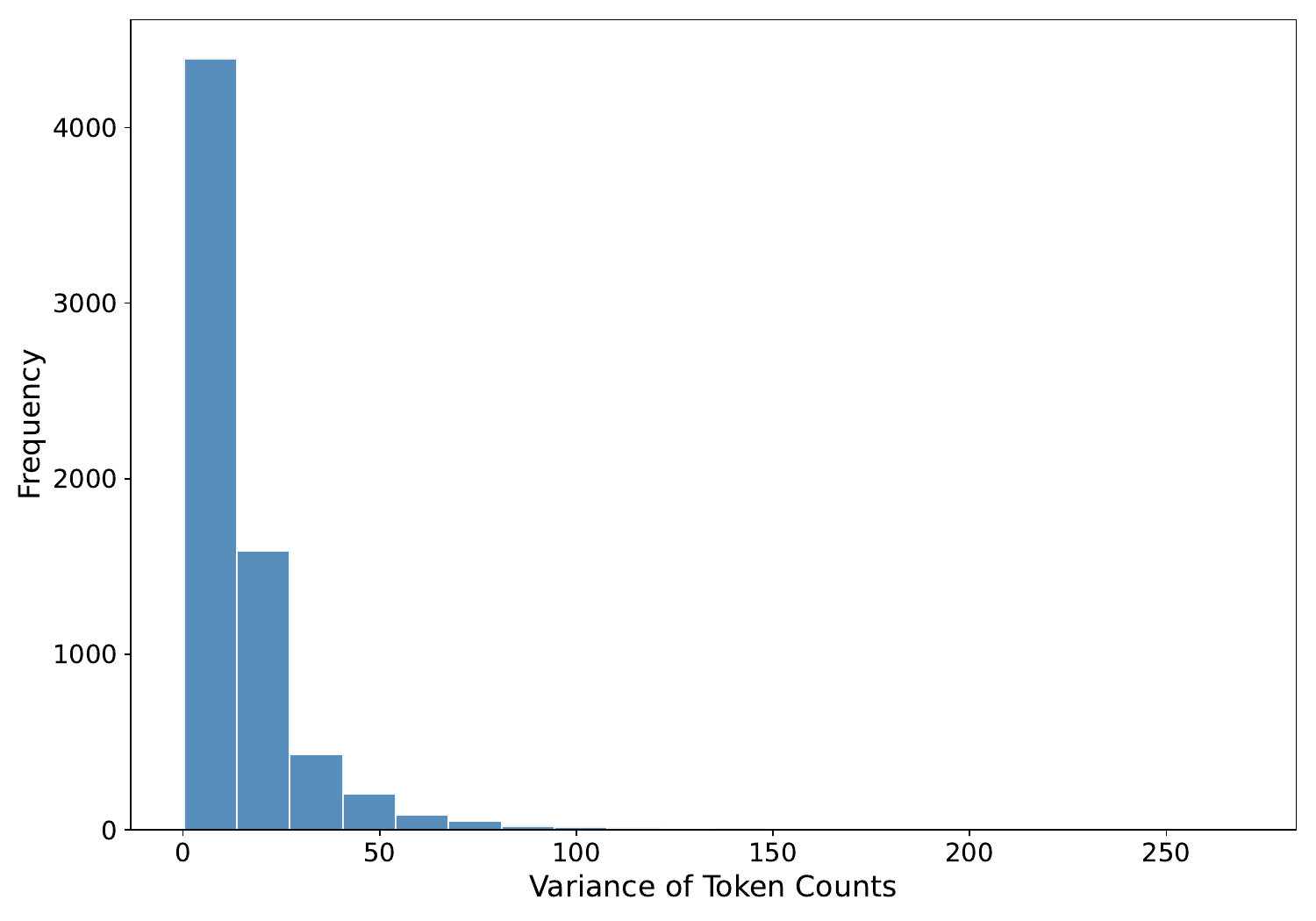}
	\caption{The variance distribution of token lengths among the six auxiliary information for the same question. The variance of most samples is close to 0.}
	\label{fig:datasetVar}
\end{figure}

Figure~\ref{fig:datasetVar} illustrates the variance distribution of token lengths among the six auxiliary information for the same question. The vast majority of items exhibit very low variance, indicating that the lengths of the six fields are balanced and comparable, thereby eliminating the interference between different answers to the same question caused by the lengths of different types of information.

\begin{figure}[h]
	\centering
	\includegraphics[width=1\linewidth]{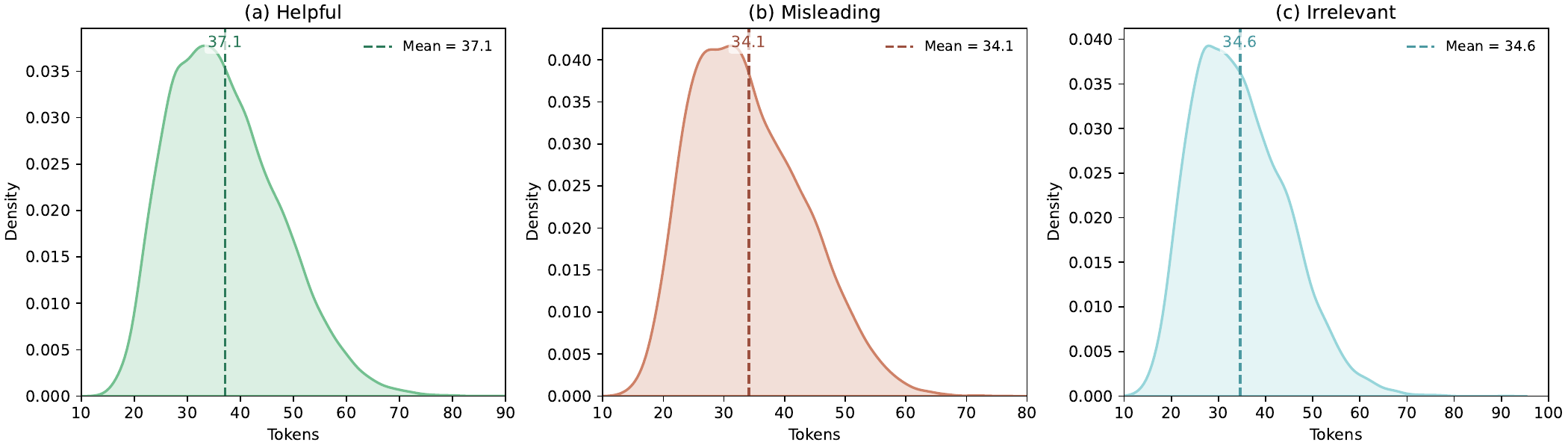}
	\caption{Token length distribution of three types of information in the \sys dataset, with the mean token length indicated by a dashed line.}
	\label{fig:avg_length}
\end{figure}

Figure~\ref{fig:avg_length} illustrates the token length distributions of three types of auxiliary information in the \sys dataset. The dashed vertical line in each subplot indicates the mean token count for that category, with Helpful averaging 37.1 tokens, Misleading averaging 34.1 tokens, and Irrelevant averaging 34.6 tokens. The distribution of token lengths of the three types of information in the figure is similar, ensuring consistency across categories and questions.

\section{Experiments}
\label{sec:design}
\subsection{Experimental Setup}
\textbf{Model.}
We use \texttt{Qwen3-8B}~\cite{yang2025qwen3} (Qwen) and \texttt{Hunyuan-7B-Instruct}\footnote{https://huggingface.co/tencent/Hunyuan-7B-Instruct} (Hunyuan) for evaluation. These models are chosen for a critical reason: its unique, explicitly designed hybrid reasoning mode. Both of them can operate in either a ``thinking'' mode for step-by-step reasoning or a ``non-thinking'' mode for rapid, intuitive responses \cite{yang2025qwen3}. This built-in duality provides an ideal testbed for a controlled comparison of different reasoning processes, which is central to our research questions. For Qwen, we use the \texttt{enable\_thinking} parameter to switch; for Hunyuan, we follow the official documentation and add \texttt{/no\_think} at the beginning of the prompt to turn off the thinking model (enabled by default).

\noindent\textbf{Task Formulation and Prompting.}
A consistent prompt template was used across all experiments to ensure comparability. The auxiliary information was presented first, followed by the question and the multiple choice options, and ended with a prompt for the answer.

\subsection{Experimental Conditions}
We conducted a comprehensive suite of experiments to isolate the effects of both information type and reasoning mode. The model's performance was evaluated under the following conditions, with each condition tested with the LRM both enabled (``Thinking ON'') and disabled (``Thinking OFF''):

$\bullet$ \textbf{Baseline:} The model is prompted only with the question and answer options, without any auxiliary information.\looseness=-1

$\bullet$ \textbf{Single Snippet:} Six auxiliary snippets are tested individually: two \textit{Helpful} (H1, H2), two \textit{Irrelevant} (I1, I2), and two \textit{Misleading} (M1, M2). 

$\bullet$ \textbf{Paired Snippets:} To simulate more complex retrieval scenarios where a model might receive multiple documents simultaneously, we further construct six pairwise combinations: H1+H2, I1+I2, M1+M2, H1+I1, H1+M1, and I1+M1.
\subsection{Evaluation Metric}
The primary evaluation metric is \textbf{accuracy}, defined as the percentage of questions for which the model correctly identified the index of the right answer among the multiple-choice options. 
\begin{figure}[h]
	\centering
	\includegraphics[width=1\linewidth]{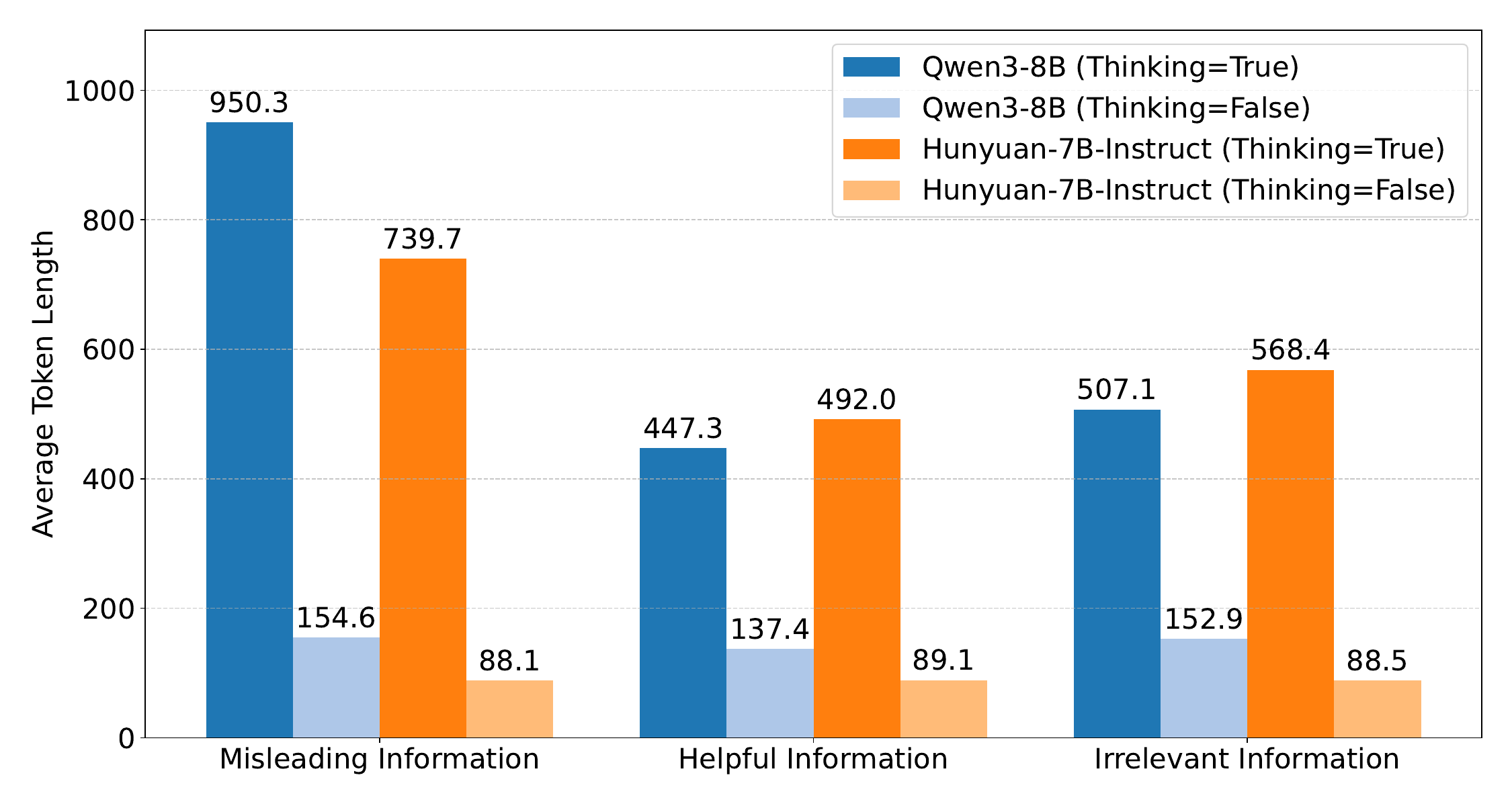}
	\caption{Average token length of responses generated by the model in different modes.}
	\label{fig:token}
\end{figure}
\subsection{Relationship between response length and input information}
Figure~\ref{fig:token} presents the average token length of responses generated by the model across different reasoning modes and types of auxiliary information. Notably, activating the explicit ``thinking mode'' consistently produces longer responses, indicating the model's propensity to generate more elaborate, step-by-step reasoning and detailed justifications when explicitly prompted.

Moreover, the length of the response is also influenced by the type of auxiliary information provided. When using thinking mode, responses generated with Helpful or Irrelevant auxiliary information are much shorter than with Misleading information. Among them, the responses with Helpful information are the shortest. We find that Helpful information helps the model reason and obtain the answer most smoothly, while Misleading information may prompt the model to elaborate further in an attempt to reconcile conflicting cues. 

These results indicate that both reasoning mode and auxiliary information quality jointly influence response verbosity. Importantly, increased response length does not necessarily correspond to greater accuracy, particularly when misleading context is present. This highlights the need to avoid equating longer or more detailed outputs with higher quality in LLM-generated responses.

\begin{table}[t]
\centering
\footnotesize
\setlength{\tabcolsep}{3.5pt}
\begin{tabular}{lccccc}
\toprule
\textbf{Info Type} 
& \textbf{OFF} & \textbf{ON} 
& $\boldsymbol{\Delta_\text{think}}$
& $\boldsymbol{\Delta_\text{base}^{\text{OFF}}}$ 
& $\boldsymbol{\Delta_\text{base}^{\text{ON}}}$ \\
\midrule
None (Baseline)      & 86.35 & 96.22 & +9.87 & -- & -- \\
\midrule
Helpful (H1)          & 90.28 & 97.06 & +6.78 & +3.93 & +0.83 \\
Helpful (H2)          & 85.88 & 94.32 & +8.44 & \cellcolor{gray!40}-0.47 & \cellcolor{gray!40}-1.90 \\
\midrule
Irrelevant (I1)        & 85.52 & 94.73 & +9.21 & \cellcolor{gray!40}-0.83 & \cellcolor{gray!40}-1.49 \\
Irrelevant (I2)        & 85.54 & 94.68 & +9.14 & \cellcolor{gray!40}-0.81 & \cellcolor{gray!40}-1.54 \\
\midrule
Misleading (M1)        & 76.20 & 79.87 & +3.67 & \cellcolor{gray!40}-10.15 & \cellcolor{gray!40}-16.35 \\
Misleading (M2)        & 74.34 & 77.30 & +2.96 & \cellcolor{gray!40}-12.01 & \cellcolor{gray!40}-18.92 \\
\midrule
Mix (H1+H2)             & 90.57 & 96.91 & +6.34 & +4.22 & +0.69 \\
Mix (I1+I2)               & 86.17 & 94.86 & +8.68 & \cellcolor{gray!40}-0.18 & \cellcolor{gray!40}-1.36 \\
Mix (M1+M2)               & 74.87 & 77.03 & +2.16 & \cellcolor{gray!40}-11.48 & \cellcolor{gray!40}-19.19 \\
Mix (H1+I1)                 & 89.46 & 97.10 & +7.64 & +3.10 & +0.88 \\
Mix (H1+M1)                 & 85.76 & 93.01 & +7.25 & \cellcolor{gray!40}-0.59 & \cellcolor{gray!40}-3.21 \\
Mix (I1+M1)                   & 78.94 & 84.31 & +5.37 & \cellcolor{gray!40}-7.41 & \cellcolor{gray!40}-11.91 \\
\bottomrule
\end{tabular}
\caption{\textbf{Qwen accuracy on \sys.} 
OFF, ON and $\Delta_\text{think}$ denotes the accuracy of w/o thinking, w/ thinking and ON$-$OFF, respectively. $\Delta_\text{base}^{\text{OFF/ON}}$ shows accuracy change from the corresponding Baseline.}
\label{tab:qwen_results}
\end{table}

\begin{table}[t]
\centering
\footnotesize
\setlength{\tabcolsep}{3.5pt}
\begin{tabular}{lccccc}
\toprule
\textbf{Info Type} 
& \textbf{OFF} & \textbf{ON} 
& $\boldsymbol{\Delta_\text{think}}$
& $\boldsymbol{\Delta_\text{base}^{\text{OFF}}}$ 
& $\boldsymbol{\Delta_\text{base}^{\text{ON}}}$ \\
\midrule
None (Baseline)        & 70.09 & 83.06 & +12.96 & -- & -- \\
\midrule
Helpful (H1)            & 81.62 & 89.29 & +7.67 & +11.53 & +6.23 \\
Helpful (H2)            & 71.43 & 84.67 & +13.24 & +1.33 & +1.61 \\
\midrule
Irrelevant (I1)             & 73.27 & 82.51 & +9.24 & +3.18 & \cellcolor{gray!40}-0.55 \\
Irrelevant (I2)             & 72.80 & 82.67 & +9.87 & +2.71 & \cellcolor{gray!40}-0.39 \\
\midrule
Misleading (M1)          & 69.70 & 71.15 & +1.45 & \cellcolor{gray!40}-0.40 & \cellcolor{gray!40}-11.91 \\
Misleading (M2)          & 63.50 & 68.61 & +5.11 & \cellcolor{gray!40}-6.59 & \cellcolor{gray!40}-14.45 \\
\midrule
Mix (H1+H2)              & 80.71 & 91.60 & +10.89 & +10.62 & +8.55 \\
Mix (I1+I2)                   & 72.99 & 82.47 & +9.48 & +2.90 & \cellcolor{gray!40}-0.58 \\
Mix (M1+M2)               & 63.59 & 62.48 & \cellcolor{gray!40}-1.11 & \cellcolor{gray!40}-6.50 & \cellcolor{gray!40}-20.58 \\
Mix (H1+I1)                         & 80.81 & 89.64 & +8.83 & +10.72 & +6.59 \\
Mix (H1+M1)                     & 74.40 & 83.26 & +8.86 & +4.31 & +0.20 \\
Mix (I1+M1)                             & 69.93 & 72.22 & +2.29 & \cellcolor{gray!40}-0.16 & \cellcolor{gray!40}-10.84 \\
\bottomrule
\end{tabular}
\caption{\textbf{Hunyuan accuracy on \sys.} The meaning of symbols is the same as in Table~\ref{tab:qwen_results}.}
\label{tab:hunyuan_results}
\end{table}

\subsection{The Quantity Effect: From Zero to Two Pieces of Information}
We first analyze how accuracy changes as the number of auxiliary snippets increases from zero to one and two, focusing on the three types of information.

As shown in Table~\ref{tab:qwen_results},\ref{tab:hunyuan_results}, \textit{for Helpful information}, adding one snippet consistently improves accuracy over the baseline for both reasoning modes.  
Providing two Helpful snippets does not lead to further improvement compared to the best single snippet; performance tends to plateau once one piece of correct information is available.

\textit{For Irrelevant information}, the accuracy differences between zero, one, and two snippets are small.  
Introducing such information causes slight fluctuations around the baseline, and combining two irrelevant snippets yields results close to those with a single snippet.

\textit{For Misleading information}, accuracy drops markedly once any misleading snippet is introduced, and the decrease becomes more pronounced when two misleading snippets are provided.  
This pattern holds in both reasoning modes.

Overall, increasing the number of auxiliary snippets produces different trends depending on their type:
Helpful information shows diminishing returns beyond one snippet, Irrelevant information shows minimal effect regardless of quantity, and Misleading information shows compounding negative effects as the number increases.

\subsection{Interaction of Information Type and Reasoning Mode}
We further examine how the effect of reasoning mode varies across information types by comparing the accuracy gap between the thinking (ON) and non-thinking (OFF) modes.

In the \textit{baseline condition}, enabling thinking substantially improves accuracy (e.g., +9.9 points in Qwen, +13.0 in Hunyuan).  
\textit{With Helpful information}, thinking still brings clear gains (about +7–8 points in Qwen and +7–13 in Hunyuan), though the margin is smaller than at baseline.  
\textit{With Irrelevant information}, the thinking gains are similar in magnitude (around +9 points in Qwen), but do not raise accuracy above the baseline level. 
\textit{With Misleading information}, the gains become very small or even negative (e.g., +2–3 points in Qwen, negative in Hunyuan for M1+M2). In particular, both models showed the largest decrease in accuracy (-18.92 and -20.58, respectively) when thinking with the two Misleading information.

Overall, the benefit of thinking mode decreases as the quality of the provided information declines:  
it is largest with no information, moderate with Helpful and Irrelevant information, and minimal or negative with Misleading information.

\subsection{Handling Conflicting Information}
Mixed conditions of different types of information are also evaluated to assess how the reasoning mode affects performance in the presence of conflicting signals.\looseness=-1

\textit{For H+I combinations}, the thinking mode produces gains similar to those with single Helpful snippets.  
For example, in Qwen the accuracy increases from 89.46\% to 97.10\%, which is close to the +7–8 point gains seen with H1 and H2.

\textit{For H+M combinations}, the benefit of thinking mode becomes smaller.  
Qwen improves from 85.76\% to 93.01\% (+7.25), which is below the Helpful gains but above the +2–3 points seen with single Misleading snippets.

\textit{For I+M combinations}, the gains are further reduced.  
Qwen rises from 78.94\% to 84.31\% (+5.37), clearly lower than the +9 points observed for Irrelevant snippets alone.

Overall, the ON–OFF gap decreases as misleading content is introduced:  
it stays high in H+I, drops in H+M, and becomes small in I+M.

\begin{tcolorbox}[colback=black!0!white, colframe=black!70, title=Insight 1]
The impact of auxiliary information on model reasoning performance is highly dependent on the type of that information. Helpful information substantially boosts performance, while Misleading information causes a catastrophic drop in accuracy. 
\end{tcolorbox}
\begin{table}[h]
\centering
\begin{tabular}{lcc}
\toprule
\textbf{Category} & \textbf{Qwen} &\textbf{Hunyuan} \\ \midrule
Deliberative Success & 10.91\% & 20.45\%  \\
Intuitive Leap & 1.05\% & 7.50\% \\
Robustly Correct & 85.31\% & 62.60\% \\
Robustly Incorrect & 2.72\% & 9.45\% \\ \bottomrule
\end{tabular}
\caption{The impact of reasoning mode on outcomes in base conditions.}
\label{tab:typology}
\end{table}
\subsection{The Double-Edged Sword of Auxiliary Information}
To further understand the intrinsic reasoning behavior of the LRM, we track the changes in each question when thinking mode is turned on and off in the baseline condition. This method isolates the effect of the reasoning process itself from the influence of external context. All the results can be divided into the following four categories, including: \textit{1. Deliberative Success (Incorrect w/o Thinking, Correct w/ Thinking),} which are the canonical "hard" problems for which the LRM is designed; \textit{2. Intuitive Leap (Correct w/o Thinking, Incorrect w/ Thinking),} which represents a counter-intuitive failure mode; \textit{3. Robustly Correct (Correct in Both Modes),} which are questions that the model finds relatively easy; \textit{4. Robustly Incorrect (Incorrect in Both Modes),} which are too difficult for the model in its current state.

The results in Table~\ref{tab:typology} quantify the complex impact of the thinking mode. 
In the base case, Deliberative Success is greater than Intuitive Leap for both models. This shows that statistically, turning on Thinking Mode may bring more benefits, but it cannot guarantee that every sample will see improvement. At the same time, this is also related to the capabilities of the model itself. Changes brought to Hunyuan by thinking are obviously greater than those brought to by Qwen.

\section{Why Does Thinking Mode Lead to Errors?}
\begin{table*}[h]
\centering
\begin{tabularx}{\textwidth}{@{} l p{3.2cm} X @{}}
\toprule
\textbf{ID} & \textbf{Type} & \textbf{Explanation}\\ \midrule
\textbf{1} & Over-analysis and Misinterpretation & The model's reasoning process becomes too granular or focuses on a secondary, less important aspect of the problem, leading it to miss the primary issue~\cite{shi2023large,yang2025llm}. \\  
\textbf{2} & Misapplication and Overgeneralization & The model correctly recalls a rule (grammatical, stylistic, etc.) but applies it incorrectly to the specific context or extends it beyond its proper scope~\cite{wei2022chain,suzgun2022challenging}. \\ 
\textbf{3} & Correct Reasoning, Flawed Execution & In these instances, the model's step-by-step logical analysis is sound and leads to the correct conclusion internally, but it fails to produce the right final answer~\cite{kadavath2022language,wang2022self,turpin2023language}. \\ 
\textbf{4} & Factual Errors and Hallucinations & The reasoning process is based on incorrect information that the model generates or recalls~\cite{ji2023survey,lewis2020retrieval}. \\ 
\textbf{5} & Indecision and Flawed Prioritization & The model identifies multiple possible lines of reasoning or interpretations but fails to prioritize the most relevant one, leading to confusion or an arbitrary guess~\cite{liu2023lost}. \\
\bottomrule
\end{tabularx}
\caption{Five types of reasons why thinking modes lead to incorrect answers.}
\label{tab:types}
\end{table*}
\begin{figure}[h]
	\centering
	\includegraphics[width=1\linewidth]{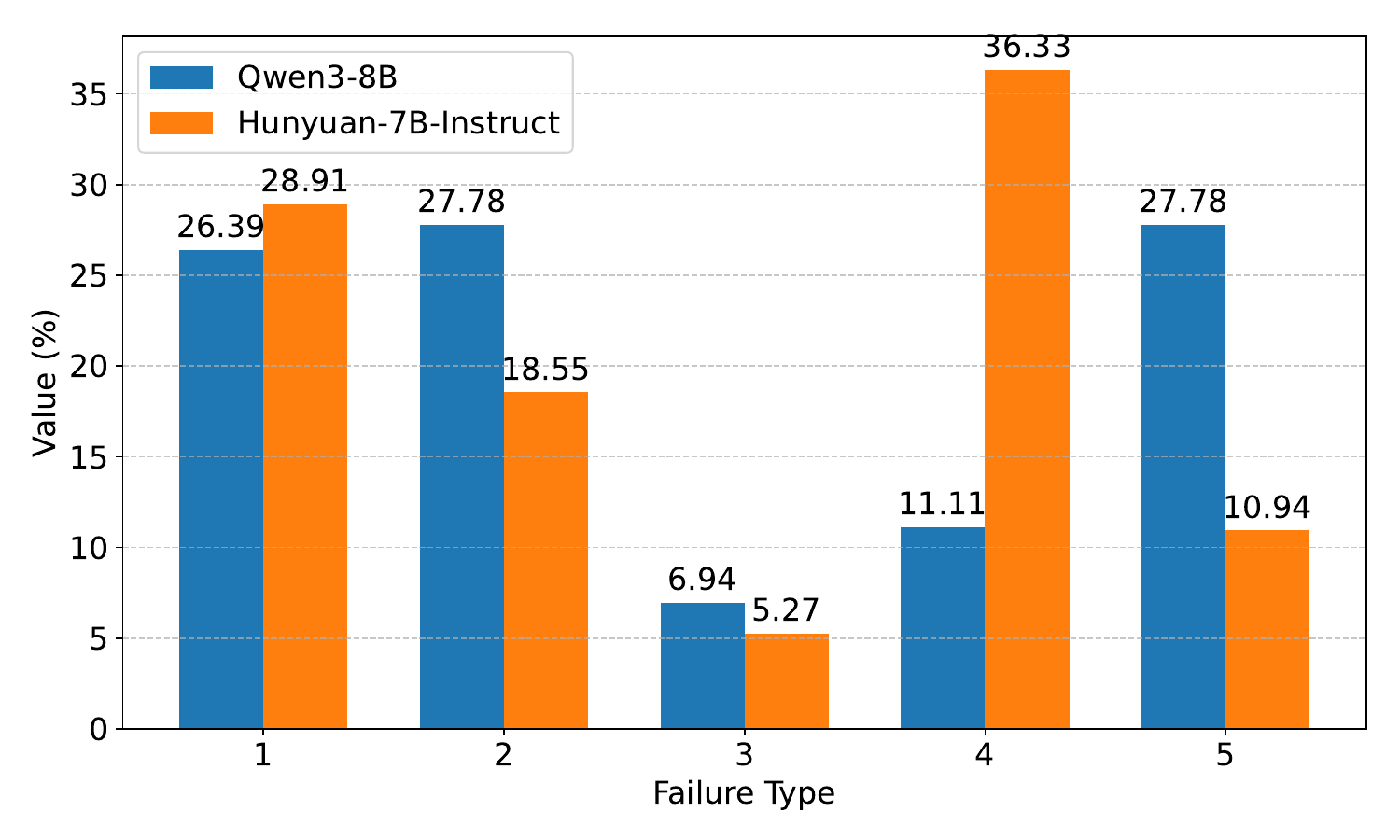}
	\caption{The proportion of incorrect answers in thinking modes caused by different reasons.}
	\label{fig:failure_type}
\end{figure}

We further focus on failure cases of the \textit{Intuitive Leap} type, which is also a direct manifestation of the side effects of thinking. The observed patterns in the LLM performance bear a striking resemblance to cognitive heuristics and biases in human psychology. We propose a classification of the observed phenomena as shown in Table~\ref{tab:types}.
It presents a systematic classification of the principal causes underlying incorrect answers produced by LLMs only when operating in ``thinking mode'':
(1) Over-analysis and Misinterpretation, (2) Misapplication and Overgeneralization, (3) Correct Reasoning, Flawed Execution, (4) Factual Errors and Hallucinations, and (5) Indecision and Flawed Prioritization. Each category reflects a distinctive failure pattern observed in the model’s reasoning traces.

Figure~\ref{fig:failure_type} presents the distribution of error types for samples that resulted in failure exclusively in thinking mode. The analysis indicates that these failures are not attributable to a single predominant cause, but are instead distributed across several categories, highlighting the multifactorial nature of reasoning breakdowns when explicit reasoning overrides an initially correct intuition. In particular, the prominence of \textit{Over-analysis and Misinterpretation} and \textit{Misapplication or Overgeneralization} across both models suggests that the deliberate reasoning process may inadvertently introduce unnecessary complexity or inappropriately apply heuristics, leading to incorrect answers. Notably, Hunyuan exhibits a higher incidence of \textit{Factual Errors and Hallucinations}, underscoring the need for model-specific strategies to address such vulnerabilities. Collectively, these findings highlight a mechanism by which explicit reasoning can undermine model performance even when the initial intuitive response is correct, providing insight for subsequent LRM training.

\section{Conclusion}
\label{sec:conclusion}
This paper provides a comprehensive evaluation of how auxiliary information shapes the reasoning capabilities of LRMs, leveraging the \sys dataset as a controlled and systematic evaluation framework. Through extensive analyses of representative LRMs, we demonstrate that the incorporation of helpful contextual information reliably enhances reasoning performance, while exposure to irrelevant or misleading auxiliary content substantially degrades accuracy. Crucially, our findings reveal that the explicit ``thinking mode'' serves as a double-edged sword: although it improves the model’s ability to address complex queries, it can also amplify vulnerability to misinformation by fostering the rationalization of erroneous or misleading premises. Collectively, these insights underscore the necessity for robust context validation mechanisms in future model architectures and highlight the importance of interdisciplinary approaches in advancing a mechanistic understanding of AI reasoning and its limitations.

\bibliography{custom}
\newpage
\appendix
\onecolumn
\section{Prompt Template}
\label{sec:apd-prompt}
The prompt template used to generate auxiliary information is shown as follows:
\begin{tcolorbox}[colback=black!0!white, colframe=black!70, title=Prompt for Generating Auxiliary Information]

I am generating information helpful, misleading, or irrelevant to questions for use in an experiment. Each data item includes the following fields:

-id: The unique identifier of the question

-question: The question that needs to be answered

-choices: The list of answer options provided in the question

-lecture: Helpful information related to the question

Based on the provided question, choices and lecture, please generate the following for each question and ouput the final result in JSON format:

-``Helpful\_information1'' and ``Helpful\_information2'': Two correct pieces of information that help answer the question. 

-``Misleading\_information1'' and ``Misleading\_information2'': Two misleading pieces of information related to the topic of the question (they may contradict the original lecture).

-``Irrelevant\_information1'' and ``Irrelevant\_information2'': Two pieces of information related to the topic, but not helpful for answering the question and not misleading either.

The following are important notes:

1. Each generated piece of information will be provided individually as background knowledge to the answering model. When there are multiple knowledge points in the options, please introduce the different knowledge points in ``Helpful\_information1'' and ``Helpful\_information2''.

2. You only need to provide **general knowledge statements** related to the subject area of the question or the choices. Don't analyze specific questions and options!

3. *Important*: Make sure that the tokens of Helpful\_information1, Helpful\_information2, Misleading\_information1, Misleading\_information2, Irrelevant\_information1, and Irrelevant\_information2 are as similar as posible !!!

Return only a JSON object with the following fields (Don't output any content other than what is specified below!):

\{

``id'': corresponding question ID,

``Helpful\_information1'': helpful information 1,

``Helpful\_information2'': helpful information 2,

``Misleading\_information1'': incorrect information 1,

``Misleading\_information2'': incorrect information 2,

``Irrelevant\_information1'': neutral unrelated information 1,

``Irrelevant\_information2'': neutral unrelated information 2

\}

Below are the data items to be processed, including id, question, choices and lecture:

\{context\}
\end{tcolorbox}

\end{document}